\documentclass[10pt,twocolumn,letterpaper]{article}

\usepackage{cvpr}              % To produce the CAMERA-READY version
\usepackage{booktabs,makecell, multirow, tabularx, amsmath, amssymb, pifont, footmisc}
\usepackage{marvosym}
\newcommand{\cmark}{\ding{51}}
\newcommand{\xmark}{\ding{55}}
\usepackage[dvipsnames]{xcolor}

\definecolor{cvprblue}{rgb}{0.21,0.49,0.74}
\usepackage[pagebackref,breaklinks,colorlinks,citecolor=cvprblue]{hyperref}

\def\paperID{7425} % *** Enter the Paper ID here
\def\confName{CVPR}
\def\confYear{2024}

\title{DanceCamera3D: 3D Camera Movement Synthesis with Music and Dance}

\vspace{-10mm}
\author{Zixuan Wang$^{1,2}$, Jia Jia$^{1,2*}$, Shikun Sun$^{1,2}$, Haozhe Wu$^{1,2}$, Rong Han$^{1}$, Zhenyu Li$^{1}$, \\
Di Tang$^{4}$, Jiaqing Zhou$^{4}$, Jiebo Luo$^{3*}$\\
\textsuperscript{\rm 1}Department of Computer Science and Technology, Tsinghua University, Beijing 100084, China\\
\textsuperscript{\rm 2}Beijing National Research Center for Information Science and Technology (BNRist)\\
\textsuperscript{\rm 3}Department of Computer Science, University of Rochester, USA \textsuperscript{\rm 4}ByteDance Hangzhou, China\\
{\tt\small \{wangzixu21, ssk21, wuhz19, hanr21, zy-li21\}@mails.tsinghua.edu.cn,  jjia@tsinghua.edu.cn,} \\
{\tt\small jluo@cs.rochester.edu, \{minliu, jiashu\}@bytedance.com}
}

\begin{document}
\twocolumn[{
\renewcommand\twocolumn[1][]{#1}
\maketitle
\vspace{-13mm}
\begin{center}
    \centering
    \includegraphics[width=\textwidth,height=8.4cm]{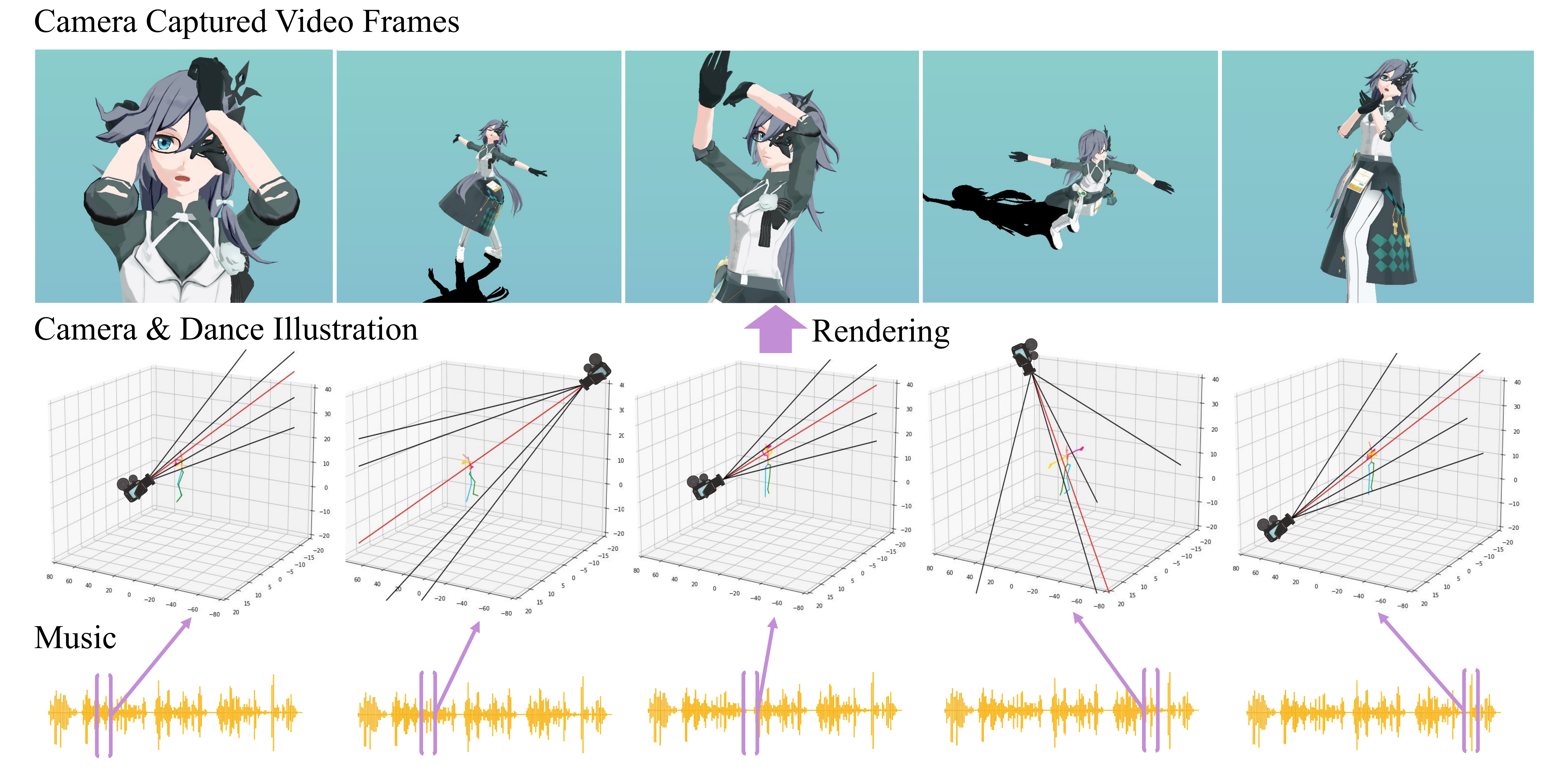}
    \vspace{-8mm}
    \captionof{figure}{We present the \textbf{DCM} dataset, which contains 3.2 hours paired 3D \textbf{D}ance motion, \textbf{C}amera movement and \textbf{M}usic audio.}
    
\end{center}

}]

% \maketitle
\newcommand\blfootnote[1]{%
\begingroup
\renewcommand\thefootnote{}\footnote{#1}%
\addtocounter{footnote}{-1}%
\endgroup
}

\blfootnote{\textsuperscript{*}~Corresponding author}
\vspace{-5mm}

\begin{abstract}
\vspace{-2mm}
Choreographers determine what the dances look like, while cameramen determine the final presentation of dances. Recently, various methods and datasets have showcased the feasibility of dance synthesis. However, camera movement synthesis with music and dance remains an unsolved challenging problem due to the scarcity of paired data. Thus, we present \textbf{DCM}, a new multi-modal 3D dataset, which for the first time combines camera movement with dance motion and music audio. This dataset encompasses 108 dance sequences (3.2 hours) of paired dance-camera-music data from the anime community, covering 4 music genres. With this dataset, we uncover that dance camera movement is multifaceted and human-centric, and possesses multiple influencing factors, making dance camera synthesis a more challenging task compared to camera or dance synthesis alone. To overcome these difficulties, we propose \textbf{DanceCamera3D}, a transformer-based diffusion model that incorporates a novel body attention loss and a condition separation strategy. For evaluation, we devise new metrics measuring camera movement quality, diversity, and dancer fidelity. Utilizing these metrics, we conduct extensive experiments on our DCM dataset, providing both quantitative and qualitative evidence showcasing the effectiveness of our DanceCamera3D model. Code and video demos are available at \url{https://github.com/Carmenw1203/DanceCamera3D-Official}.
\vspace{-6mm}
\end{abstract}

\vspace{-2mm}
\section{Introduction}
\label{sec:intro}
Dancing with the camera is a unique cinematic experience that merges cinematography and choreography. In the process of dance video production, moving the camera along with the dancer better captures the focus of dance motions and provides the audience with a more immersive storytelling experience. However, dance camera movement is influenced by multifaceted factors including music and dance. Going further, a good dance camera should have diverse shot-type changes and human-centric characteristics. As a result, creating camera movement for dance is uniquely challenging. Meanwhile, existing methods~\cite{li2021ai, siyao2022bailando,tseng2023edge} usually produce dance videos without camera movement which results in a boring single fixed view for the audience and uncontrollable situations where the dancer moves out of sight. Thus, how to automatically synthesize associated camera movement given music and dance is a significant and worth studying question.

Extensive works have made progress in music-dance dataset construction and synthesizing dance conditioned on music. However, camera movement generation with music and dance remains an open problem. This is mainly due to the following two main challenges:

\begin{itemize}[leftmargin=*]
\item[--]{\bfseries Lack of Dance Camera Data.} Previous music-dance datasets acquire 2D data from dance videos or collect 3D data using the following three methods: motion capture (MoCap), 2D to 3D reconstruction, and animator edit. However, all previous datasets only focus on music and dance, or have difficulties capturing the mobile camera pose and trajectory. Specifically, MoCap-based methods~\cite{tang2018dance, zhuang2022music2dance, valle2021transflower, alemi2017groovenet} rely on multi-camera with fixed positions or inertial sensors to achieve better accuracy so that it's harder and more complicated to use another mobile camera in the system and record the related parameters.  Reconstruction-based methods~\cite{moltisanti2022brace,yao2023dance,le2023music,li2021ai,sun2020deepdance} have problems extracting camera movements from dance video since it's confusing for the model to distinguish between camera movement and dancer movement. Animator-edited methods are suitable for camera movement design but previous related datasets~\cite{li2022danceformer, chen2021choreomaster}  only collect music audio and dance motion. In addition, some previous methods explore the camera movement extraction from 2D films, however relative positions of two characters are needed which is a very specific condition and cannot be applied to dance situations.
\item[--]{\bfseries Complexity of Dance Cinematography.} Unlike dance choreography and normal cinematography, dance cinematography considers 1) multifaceted representation of camera movement including trajectory, direction, and field of view, 2) human-centric features denoting shot types and changes such as long shot, medium shot, close-up, cut-in, and cut-out, and 3) correlation with music and dance which indicates different moving speed, shot types, and body parts attention according to music and dance. Therefore, dance camera synthesis is more complicated than dance synthesis or normal cinematography.
\end{itemize}

To address the above issues, in this paper, we construct \textbf{DCM}, a new multi-model 3D \textbf{D}ance-\textbf{C}amera-\textbf{M}usic dataset, which for the first time collects camera keyframes and movements along with music and dance to advance the study of dance cinematography. We collect 108 dance sequences of paired dance-camera-music data from the anime community, which sum up to 3.2 hours and cover 4 languages of music. 

With this dataset, we present \textbf{DanceCamera3D}, a transformer-based diffusion network, the first model that can robustly synthesize camera movement given music and dance. To better balance the effect of music and dance motion to camera movement, we propose a strong-weak condition separation strategy for classifier-free guidance (CFG)~\cite{ho2022classifier}. Meanwhile, we devise a new body attention loss to help DanceCamera3D achieve better focus on different limb parts. In addition, we introduce new metrics considering shot features and fidelity to the dancing character which are significant in dance cinematography. Using these new metrics and some rational metrics from dance synthesis, we conduct comprehensive quantitative and qualitative evaluations on our \textbf{DCM} dataset, which demonstrate that our DanceCamera3D outperforms the baseline models on quality, diversity, and fidelity. Experiments also approve that our strong-weak condition separation strategy helps the diffusion model acquire more feasibility in the trade-off among quality, diversity, and dancer fidelity. In summary, our contributions are as follows:
\begin{itemize}
\item We construct a new \textbf{DCM} dataset, which for the first time collects rich annotated camera data with multi-genre music and dance. Our DCM dataset possesses the potential to benefit the studies of dance camera synthesis, camera keyframing, and shot type classification.
\item We introduce a novel Music Dance driven Camera Movement Synthesis task, which aims to automatically synthesize camera movement given music and dance. To our best knowledge, this is the first work that proposes and works on such a problem. To conduct comprehensive evaluations, we devise new metrics considering dance cinematography knowledge.
\item We present \textbf{DanceCamera3D}, a transformer-based diffusion model, which is the first model for camera synthesis from music and dance. DanceCamera3D achieves more feasibility and better fidelity with a strong-weak condition separation strategy and a novel loss function.
\end{itemize}

\section{Related Work}
\label{sec:relatedwork}
\begin{table*}[h]
\small
\setlength{\belowcaptionskip}{0.cm}
\newcommand{\tabincell}[2]{\begin{tabular}{@{}#1@{}}#2\end{tabular}}
  \begin{tabular}{cccccccc}
    \toprule{}
    \multirow{2}*{\tabincell{c}{Dataset}} &  \multirow{2}*{\tabincell{c}{Camera \\ Data}} & \multirow{2}*{\tabincell{c}{Camera \\ Keyframes}} & \multirow{2}*{\tabincell{c}{Camera Data \\ Acquisition}}&\multirow{2}*{\tabincell{c}{Dance \\ Data}} & \multirow{2}*{\tabincell{c}{Dance \\ Keyframes}}  & \multirow{2}*{\tabincell{c}{Dance Data \\ Acquisition}}  & \multirow{2}*{\tabincell{c}{Capture \\ Environment}}\\
    \\
    \midrule 
    AIST~\cite{tsuchida2019aist} & Fixed-multi-view & \xmark & 2D videos & 2D & \xmark & 2D videos & Lab-control \\
    GrooveNet~\cite{alemi2017groovenet} & \xmark & \xmark& \xmark & 3D & \xmark & MoCap & Lab-control\\
    Dance with Melody~\cite{tang2018dance} & \xmark & \xmark& \xmark & 3D & \xmark & MoCap & Lab-control\\
    FineDance~\cite{li2023finedance} & \xmark & \xmark& \xmark & 3D & \xmark & MoCap& Lab-control \\
    AIST++~\cite{li2021ai} & \xmark & \xmark& \xmark & 3D & \xmark & Reconstruction & Lab-control\\
    AIST-M~\cite{yao2023dance} & \xmark & \xmark& \xmark & 3D & \xmark & Reconstruction & Lab-control\\
    AIOZ-GDANCE~\cite{le2023music} & \xmark & \xmark& \xmark & 3D & \xmark & Reconstruction & Lab-control\\
    ChoreoMaster~\cite{chen2021choreomaster} & \xmark& \xmark & \xmark & 3D & \xmark & Animator & In-the-wild\\
    PhantomDance~\cite{li2022danceformer} & \xmark& \xmark & \xmark & 3D & \cmark & Animator& In-the-wild\\
    Jiang \emph{et al}.~\cite{jiang2020example}& Movable-view & \xmark & Reconstruction & \xmark & \xmark & \xmark& In-the-wild\\
    Bonatti \emph{et al}.~\cite{bonatti2021batteries}& Movable-view & \xmark & Defined-Rules & \xmark & \xmark & \xmark& Lab-control\\
    Yu \emph{et al}.~\cite{yu2022bridging}& Fixed-multi-view & \xmark & Animator & \xmark & \xmark & \xmark& Lab-control\\
    \textbf{DCM} & Movable-view & \cmark& Animator & 3D & \cmark & Animator& In-the-wild\\
  \bottomrule
\end{tabular}
\caption{\textbf{Comparsion of dance-camera-music datasets}. Our DCM dataset is the first 3D dataset with dance, music, and camera movement including keyframe data, which can benefit the studies of dance camera synthesis, camera keyframing, and shot type classification.}
\vspace{-5mm}
\label{tab:dataset-comparsion}
\end{table*}

\subsection{Dance and Camera Dataset}
The construction of 2D and 3D music-dance datasets has attracted much attention since data-driven methods became popular in dance synthesis. Early works construct 2D music-dance datasets from videos. Authors of~\cite{lee2019dancing} extract 2D skeleton from dance videos using 2D pose estimation~\cite{8765346}. AIST~\cite{tsuchida2019aist} provides multi-view dancing videos paired with music. Meanwhile, tremendous progress has been made in the construction of 3D dance datasets. From the perspective of data acquisition methods, 3D dance datasets can be divided into three categories: motion capture (MoCap) based, reconstruction-based, and animator-edited datasets. Authors of~\cite{tang2018dance, zhuang2022music2dance, valle2021transflower, alemi2017groovenet, li2023finedance} utilize MoCap to record 3D skeletons to build music-dance datasets. Considering the high cost for hiring dancers and equip devices of MoCap system, authors of~\cite{moltisanti2022brace,yao2023dance,le2023music,li2021ai,sun2020deepdance, wang2022groupdancer} propose to extract 3D dance pose from 2D dance video with tracking and pose estimation tools like AlphaPose~\cite{alphapose} and MMPose~\cite{mmpose2020}. Unlike the above two types of 3D datasets, animator-edited datasets~\cite{li2022danceformer, chen2021choreomaster} are built from anime communities or hiring animators to annotate dance motions aligned with music. However, previous music-dance datasets all focus on music and dance data acquirement or have problems in capturing movable camera movement. Therefore, the camera correlation with music and dance is not exploited in these datasets. Besides, camera datasets are constructed for camera planning studies. Specifically, authors of~\cite{jiang2020example} extract camera movement from 2D films, which relies on the positions of two characters. Authors of~\cite{bonatti2021batteries} produce multi-view tracking camera data with pre-defined movement rules in a photo-realistic simulator. Authors of~\cite{yu2022bridging} manually edit multi-camera with fixed positions for the study of camera placement in storytelling situations. Overall, existing camera datasets are limited in pre-defined rules or 2D estimation methods which have many constraints. Yet, movable camera data in dance situations is ignored. We compare different music-dance datasets in Table \ref{tab:dataset-comparsion}.

\subsection{Dance Synthesis}
Dance synthesis and dance camera synthesis are closely related problems since they share significant procedures including music feature extraction, encoding of dance motions and music features, and spatio-temporal forecasting. Extensive works have made progress in dance synthesis. Early methods~\cite{cardle2002music,lee2013music, ofli2011learn2dance,fan2011example} regard music-to-dance as a similarity-based or statistical retrieval problem, which results in unrealistic choreographies and limited capacity. With the development of deep learning methods and large-scale datasets, researchers utilize neural networks to solve this problem. Typically, Crnkovic-Friis \emph{et al}.~\cite{crnkovic2016generative} devise a Chor-RNN framework to predict dance motion. Tang \emph{et al}.~\cite{tang2018dance} synthesize dance motion using an LSTM-autoencoder. Wu \emph{et al}.~\cite{wu2021dual} employ Generative Adversarial Networks (GANs) to learn both music-to-dance and dance-to-music. Later, authors of~\cite{li2021ai, siyao2022bailando, siyao2023bailando++} propose transformer-based methods auto-regressively, and authors of~\cite{tseng2023edge} use a diffusion model to synthesize dance in a denoising way. Meanwhile, some previous works~\cite{ye2020choreonet, chen2021choreomaster} introduce motion units from dance knowledge to produce more realistic dance, and some others~\cite{wang2022groupdancer,le2023music,yao2023dance} make efforts to generate group dance. However, previous approaches all focus on dance synthesis and ignore the significance of synthesizing camera movement in dance performance.
\subsection{Camera Control and Planning}
Automatic cinematography has attracted growing interest since manually producing film-like videos needs both professional practice and high labor but artistic video content is crucial in media entertainment and game industry. Jiang \emph{et al}.~\cite{jiang2020example} propose to extract camera behaviors from film clips and re-apply these behaviors in a virtual environment. Rao \emph{et al}.~\cite{rao2023dynamic} take story and camera scripts as input to compose dynamic storyboards with changing camera views in a virtual environment. Wu \emph{et al}.~\cite{wu2023secret} propose a GAN-based controller which generates actor-driven camera movements considering spatial, emotional, and aesthetic factors. Rucks \emph{et al}.~\cite{rucks2021camerai} present CamerAI to imitate chase camera in third-person games which have viewpoints constraints. To produce better cutscenes in games, Evin \emph{et al}.~\cite{evin2022cine} devise Cine-AI by mimicking the cinematography techniques of movie directors. In addition, authors of~\cite{huang2019learning,8793915,9381626,8967592} make efforts in aerial cinematography studies which aim to automatically generate movement of camera drone with artistic principles and film style. Compared to previous problems, camera control and planning in dance is more challenging because the correlation of the camera movement with music and dance motions should be considered.
\section{The DCM Dataset}
\label{sec:dataset}

\subsection{Dataset Collection and Preprocessing} 
Since MoCap and reconstruction methods have difficulties capturing camera movement along with dance motion, we collect paired dance-camera-music data from the anime community. The raw data is MikuMikuDance (MMD) resources in which dance motions and camera movements are represented as keyframes with Bezier Curve parameters. However, the Bezier Curve makes it a non-differentiable process to calculate the in-between frames which is not suitable for back-propagation. Thus, for training with neural networks, we calculate motions for each joint and interpolate frames between keyframes with Bezier interpolation. Afterward, we align the dance, camera, and music data.

\subsection{Dataset Description} \label{sec:dataset_description}
\begin{figure}[h]
  \includegraphics[width=\linewidth]{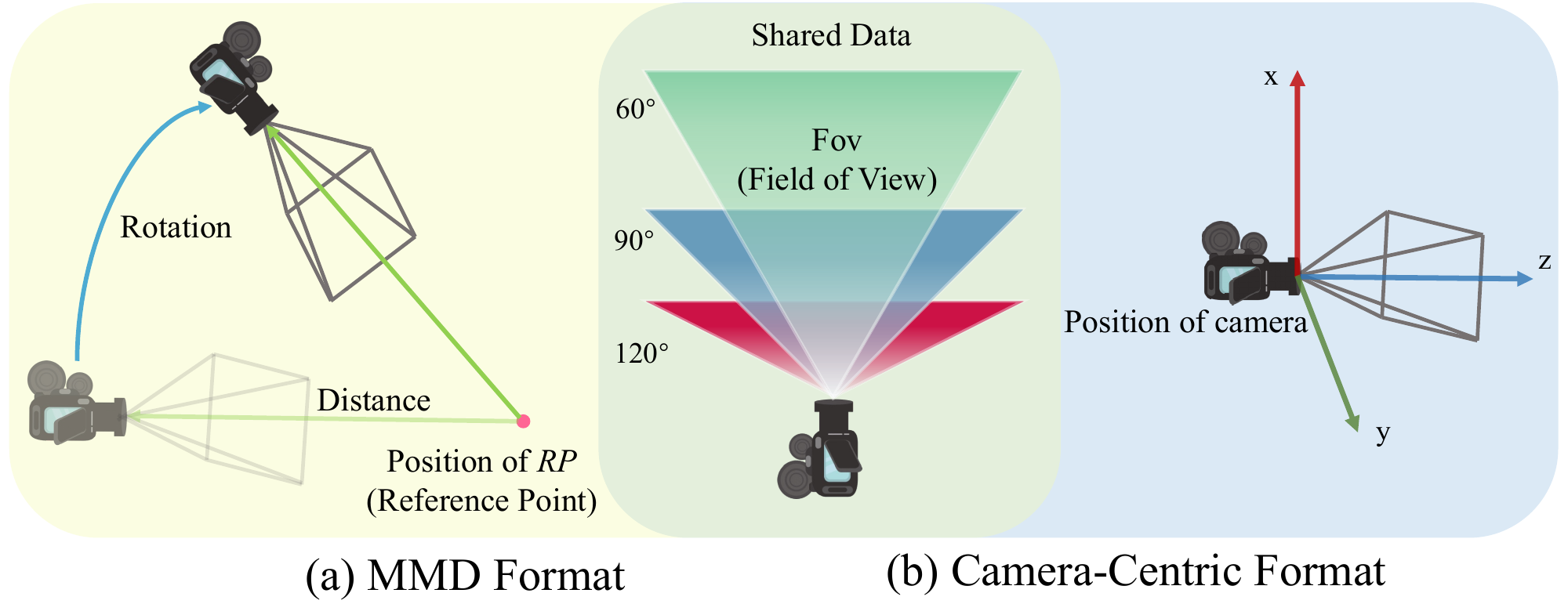}
  \caption{\textbf{Camera pose formats in our DCM dataset.} (a) shows the original MMD format of camera pose including the position of RP, rotation and distance relative to RP, and Fov. (b) illustrates our Camera-Centric format consisting of the camera's Fov, global position, and rotation represented with x, y, and z vectors in the above figure.
  }
  \label{fig:camera_representation}
\end{figure}

After alignment, we have 108 pieces of paired data(193 minutes) covering music in 4 kinds of languages including English, Chinese, Korean, and Japanese. For camera pose representation, we originally acquired data in the MMD format. As shown in (a) of Figure~\ref{fig:camera_representation}, we assume RP is the reference point of camera pose, then the MMD format camera data includes the position of RP, rotation and distance relative to RP, and camera Fov(field of view). This format is not enough for training because it cannot directly reflect the spatial relation between camera and character, and absolute camera trajectory which are significant to dance camera synthesis. Thus, we calculate a Camera-Centric format which consists of global position, rotation, and Fov of the camera, as shown in (b) of Figure~\ref{fig:camera_representation}. Besides, dance motion in our data consists of rotations and positions of 60 joints. The FPS of dance motion and camera movement is 30.

\begin{figure}[h]
  \includegraphics[width=1.05\linewidth]{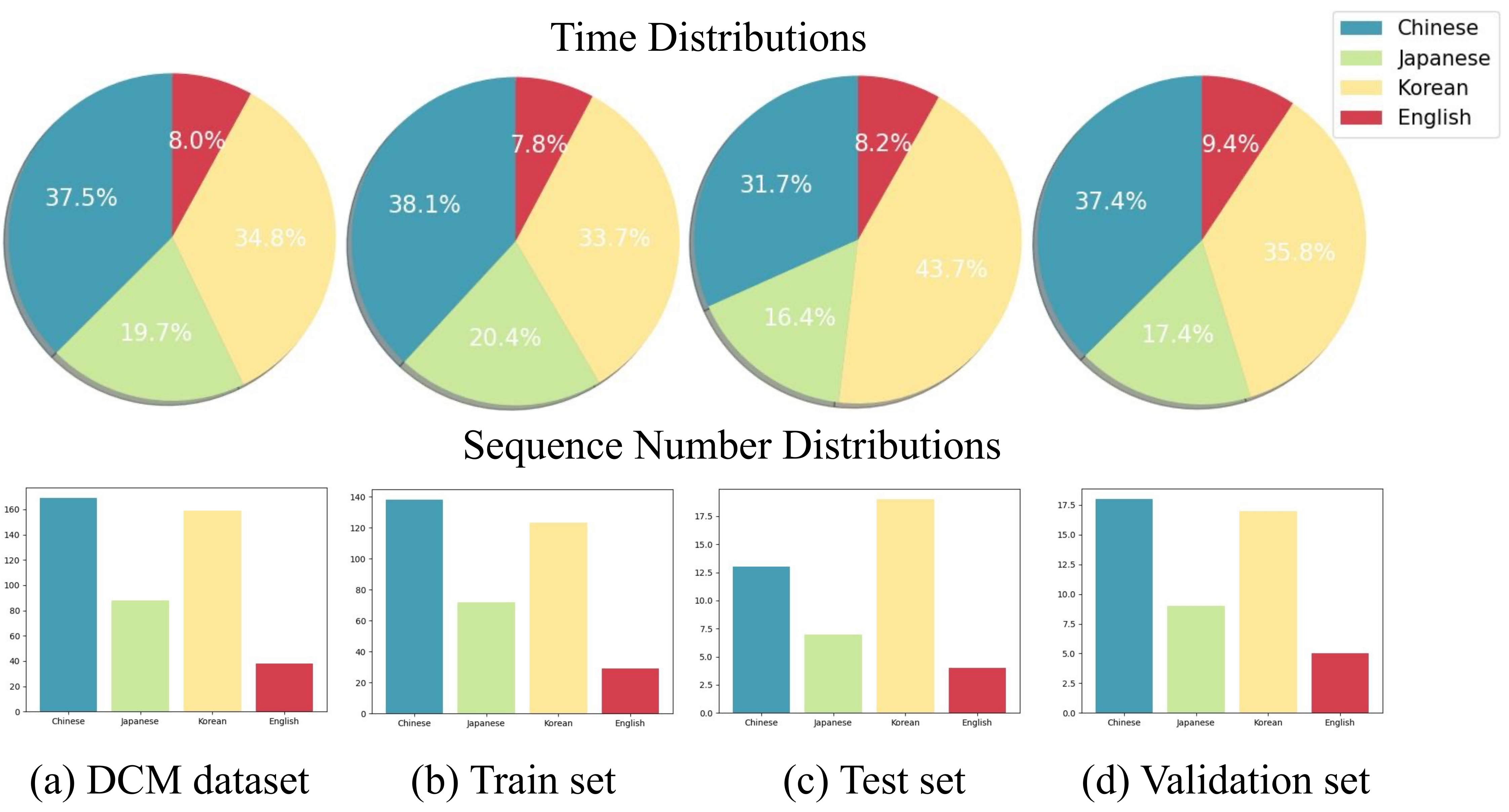}
  \caption{\textbf{Detailed distributions of our DCM dataset and split sets.}
  }
  \label{fig:dataset_pies}
  \vspace{-5mm}
\end{figure}

\subsection{Dataset Split}\label{sec:datasplit}
The duration of the original data ranges from 17 to 267 seconds, so simply dividing them into the train, test, and validation sets cannot take both the music types and duration into account. To solve this problem, we first randomly cut our data into shorter pieces ranging from 17 to 35 seconds, in which all cut points are keyframes for better reservation of camera characteristics. Then for every music type, we randomly split the data with probabilities of 0.8 : 0.1 : 0.1 to obtain the train, test, and validation sets. As shown in Figure~\ref{fig:dataset_pies}, we illustrate the detailed distributions of the split sets and our whole dataset.

\section{Music \& Dance Driven Camera Generation}
\label{sec:problemmethod}
\begin{figure*}[h]
  \includegraphics[width=1\textwidth]{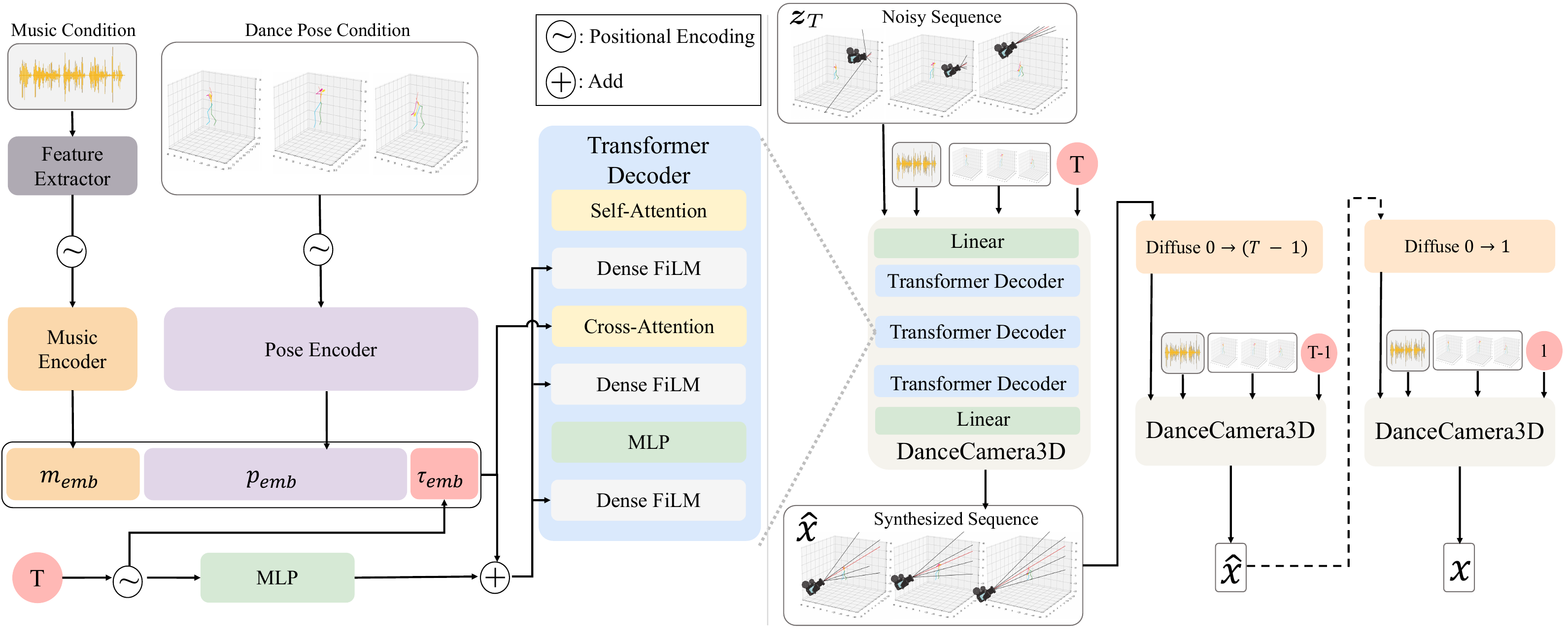}
  \caption{\textbf{Overview of DanceCamera3D Framework.} We adopt a transformer-based diffusion architecture to synthesize dance camera movement given music audio and dance pose as conditions. DanceCamera3D takes above conditions and a noisy sequence $\boldsymbol{z}_{T} \sim \mathcal{N}(0,\boldsymbol{I})$ as input and predicts noiseless sample $\hat{\boldsymbol{x}}$. Then we diffuse back $\hat{\boldsymbol{x}}$ and repeat the denoising process until $t=0$ to acquire final results.}
  \label{fig:framework}
\end{figure*}
\begin{figure}[h]
  \includegraphics[width=\linewidth]{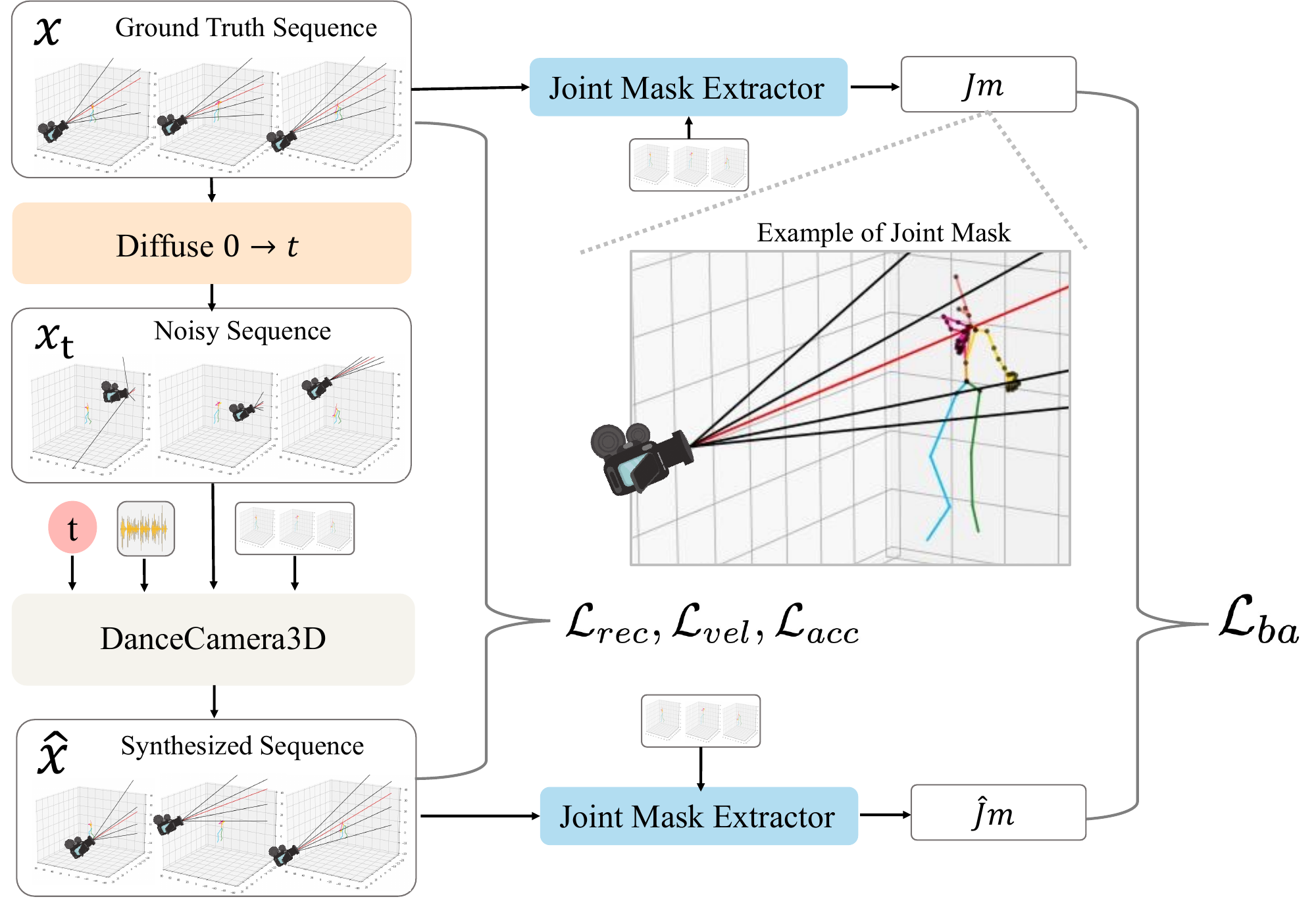}
  \caption{\textbf{Illustration of the training process and losses.} For each randomly sampled timestep $t$, we diffuse back the ground truth sequence to a noisy sequence. Then DanceCamera3D takes conditions, timestep, and a noisy sequence to predict camera movements $\hat{\boldsymbol{x}}$. We propose to detect joint masks indicating joints inside the camera view and devise the body attention loss $\mathcal{L}_{ba}$ based on joint masks which are represented with dots on the joints.}
  \label{fig:train_loss}
\end{figure}
\subsection{Problem Formulation}
The problem setting of music and dance conditioned camera generation is to predict the movement of the camera from given aligned music audio and dance poses. Here we represent music audio and dance pose conditions as $\boldsymbol{m} = \{m^1,m^2,…,m^N\}$ and $\boldsymbol{p} = \{p^1,p^2,…,p^N\}$ for a sequence with $N$ frames. Since dance motion data in MMD resources have 60 frequently used joints, we represent dance pose with joints global positions: $p^i \in \mathbb{R}^{{60}\times{3}}$. For the camera representation, we use the MMD format: $\boldsymbol{x} = \{x^1,x^2,…,x^N\},$ $x^i\in\mathbb{R}^{{3+3+1+1}}$ for training, and calculate the camera-centric format: $\boldsymbol{xc} = \{xc^1,xc^2,…,xc^N\},$ $xc^i\in\mathbb{R}^{{3+3*3+1}}$ for some loss functions. Overall, DanceCamera3D learns to predict camera $\boldsymbol{x}$ from input music $\boldsymbol{m}$ and dance $\boldsymbol{p}$.
\subsection{DanceCamera3D Architecture}
As shown in Figure~\ref{fig:framework}, DanceCamera3D uses a transformer-based diffusion model to synthesize camera movement in a denoising manner. Given music and dance pose conditions $\boldsymbol{m}$ and $\boldsymbol{p}$, we first extract the acoustic feature and then encode $\boldsymbol{m}$ and $\boldsymbol{p}$ to obtain music and pose embeddings $\boldsymbol{m}_{emb}$ and $\boldsymbol{p}_{emb}$. Then for generation process, we follow the DDPM~\cite{ho2020denoising} to define the diffusion as a Markov noising process with T steps, in which latents {$\boldsymbol{x}_t$}$^T_{t=0}$ follow a forward noising process $q(\boldsymbol{x}_t|\boldsymbol{x})$:
\begin{equation}
q(\boldsymbol{x}_t|\boldsymbol{x}) \sim \mathcal{N}(\sqrt{\bar{\alpha}_t}\boldsymbol{x},(1-\bar{\alpha}_t)\boldsymbol{I}),
\end{equation}
where $\boldsymbol{x} \sim p(\boldsymbol{x})$ is sampled from data distribution and $\bar{\alpha}_t \in (0,1)$ are monotonically decreasing constants. In this way, we can approximately produce $\boldsymbol{x}_{T} \sim \mathcal{N}(0,\boldsymbol{I})$ when $\bar{\alpha}_t$ approaches 0. Reversely, our model learns to predict $\hat{\boldsymbol{x}}(\boldsymbol{x}_t, t, \boldsymbol{m}, \boldsymbol{p}) \approx \boldsymbol{x}$ for all $t$. Thus, our DanceCamera3D takes music, dance and a noisy sequence $\boldsymbol{z}_{T} \sim \mathcal{N}(0,\boldsymbol{I})$ as input to predict noiseless camera movement $\hat{\boldsymbol{x}}$. For inference, we noise $\hat{\boldsymbol{x}}$ back to timestep $t-1$ as $\boldsymbol{x}_{t-1}$ and repeat the denoising process until $t=0$ to obtain final results.

\subsection{Training and Losses}
We illustrate the train process and losses for DanceCamera3D in Figure~\ref{fig:train_loss}. Each time, we first randomly sample $t \in (0,T)$ and $\boldsymbol{x}$ from ground truth distribution. Then add noise for $\boldsymbol{x}$ to $\boldsymbol{x}_t$ with $q(\boldsymbol{x}_t|\boldsymbol{x})$. Afterward, we enter $\boldsymbol{m}$, $\boldsymbol{p}$, and $t$ into the model and acquire synthesized camera movement sequence $\hat{\boldsymbol{x}}$. We train our model with commonly used classifier-free guidance (CFG)~\cite{ho2022classifier} in diffusion models. However, considering music and dance have quite different impacts on camera movement, we propose a strong-weak condition separation strategy to conduct CFG on these two conditions respectively instead of together, which is demonstrated to be effective in Sec~\ref{sec:cmpcfg}.
So far, we can restrict the synthesized results to comply with the conditions using losses. For physical realism, we select commonly used reconstruction loss $\mathcal{L}_{rec}$, velocity loss $\mathcal{L}_{vel}$ and acceleration loss $\mathcal{L}_{acc}$:
\begin{equation}
\begin{split}
\mathcal{L}_{rec} &= ||\boldsymbol{x} - \hat{\boldsymbol{x}}||_2^2, \\
\mathcal{L}_{vel} &= ||\boldsymbol{x}' - \hat{\boldsymbol{x}}'||_2^2, \\
\mathcal{L}_{acc} &= ||\boldsymbol{x}'' - \hat{\boldsymbol{x}}''||_2^2,
\end{split}
\end{equation}
where $\boldsymbol{x}'$ and $\boldsymbol{x}''$ represent the first-order (velocity) and second-order (acceleration) partial derivatives of camera movement parameters $\boldsymbol{x}$ on time. However, these commonly used losses for movement synthesis cannot help the model to capture the significance of the dancer's motion and even move the dancer out of camera view, for which we provide a more detailed discussion in Sec~\ref{sec:ablation}. To solve this problem, we propose a body attention loss $\mathcal{L}_{ba}$:
\begin{equation}
\mathcal{L}_{ba} = ||\boldsymbol{Jm} - \boldsymbol{\hat{J}m}*\boldsymbol{Jm}||, 
\end{equation}
where $\boldsymbol{Jm}$ denotes whether joints are inside or outside the camera view:
\begin{equation}
\boldsymbol{Jm}_j^{i} = 
\begin{cases}
1& p_j^i  \enspace \text{inside Camera View}, \\
0& p_j^i  \enspace \text{outside Camera View},
\end{cases}
\end{equation}
where $p_j^i$ refers to position of joint $j$ at frame $i$. 
For better visualization of the bone mask, we show a sample with joint dots in Figure~\ref{fig:train_loss}. More details on the implementation of $\mathcal{L}_{ba}$ and $\boldsymbol{Jm}$ are illustrated in supplementary materials.
Using $\mathcal{L}_{ba}$, the model is restricted to concentrate more on significant body parts that are captured in ground truth.
Our overall training loss consists of a weighted sum of the above losses, while $\lambda_{vel}, \lambda_{acc}, \lambda_{ba}$ are trade-off weights:
\begin{equation}
\mathcal{L} = \mathcal{L}_{rec} + \lambda_{vel}\mathcal{L}_{vel} + \lambda_{acc}\mathcal{L}_{acc} + \lambda_{ba}\mathcal{L}_{ba}. 
\end{equation}

\section{Experiments}
\label{sec:experiments}
\begin{table*}[h]
\setlength\tabcolsep{6.5pt}% 调整列间距
\vspace{-2mm}
  \begin{tabular}{lccccccc}
    \toprule{}
    \multirow{2}{*}[-0.5ex]{Method}& \multicolumn{2}{c}{Quality} &\multicolumn{2}{c}{Diversity}&\multicolumn{2}{c}{Dancer Fidelity}&User Study\\
    \cmidrule(lr){2-3} \cmidrule(lr){4-5} \cmidrule(lr){6-7} \cmidrule(lr){8-8}
    &FID$_k\downarrow$ & FID$_s\downarrow$& Dist$_k\uparrow$  & Dist$_s\uparrow$ &DMR$\downarrow$ & LCD$\downarrow$ & DanceCamera3D WinRate$\uparrow$\\
    \midrule
    Ground Truth  & - & - &  3.275 &1.731 &  0.00142 &-& 37.62\% $\pm$ 2.83\%\\
    \midrule
   DanceRevolution$^{*}$~\cite{huang2021} & 10.267 & 2.368&1.491 & 1.118&0.0062& 0.154&71.90\% $\pm$ 2.38\%\\
    FACT$^{\#}$~\cite{li2021ai}  & 5.205 & 0.960 & 1.505 & 1.007&0.0070&0.151&65.71\% $\pm$ 1.71\%\\
    DanceCamera3D w/o $\mathcal{L}_{ba}$  & 4.022& 0.728 & 1.421  & $\mathbf{1.671}$ & 0.0899 &0.310& 77.14\% $\pm$ 3.53\%\\
    DanceCamera3D (Ours) & $\mathbf{3.749}$  & $\mathbf{0.280}$& $\mathbf{1.631}$ & 1.326 & $\mathbf{0.0025}$ & $\mathbf{0.147}$&-\\
  \bottomrule{}
  
\end{tabular}
\vspace{-5mm}
\caption{\textbf{Comparison of our DanceCamera3D with Spatio-Temporal Models.} $^{*}$ means we utilize the LSTM decoder of DanceRevolution~\cite{huang2021} to generate camera motion frame by frame. $^{\#}$ means we follow FACT~\cite{li2021ai} to autoregressively synthesize camera motion with seed motions. - denotes that the self-comparison is meaningless.}
\label{tab:comparsononbackbone}
\vspace{-5mm}
\end{table*}
\subsection{Experimental Setup}
\textbf{Dataset Preparation.} All the experiments in this paper are conducted on our DCM dataset which for the first time collects camera movement with dance and music. As mentioned in Sec~\ref{sec:datasplit}, we split DCM into train, test, and validation sets, and report the performance on the test set only. For the training of DanceCamera3D, we split the train data into 5-second subsequences with a stride of 0.5 seconds.\\
\textbf{Implementation Details.} In our experiment, the input of the model is aligned dance motion and music audio except for a transformer baseline needs an extra 2.5-second (75 frames) camera seed movements. The output of the model is camera movement sequences with the same length to input dance and music. During inference, we generate 5-second subsequences with a stride of 2.5 seconds, then we interpolate the overlapping slices to enforce consistency with linear decaying weight. Afterward, we use a total variation denoiser~\cite{du2018minmax} to detect the keyframes in our results and use a Savitzky-Golay filter~\cite{schafer2011savitzky} to smooth camera movements between keyframes. For the training process, all our experiments are trained with 128 batch size for 3000 epochs using Adan~\cite{xie2022adan} optimizer. Our final model has 52.7M total parameters, which was trained on 6 NVIDIA 3090 GPUs for 13 hours. We utilize “Jukebox” to extract music features that have 4800 dimensions information for each frame. For diffusion timesteps, we use $T = 1000$.
\begin{figure}[h]
  \includegraphics[width=\linewidth]{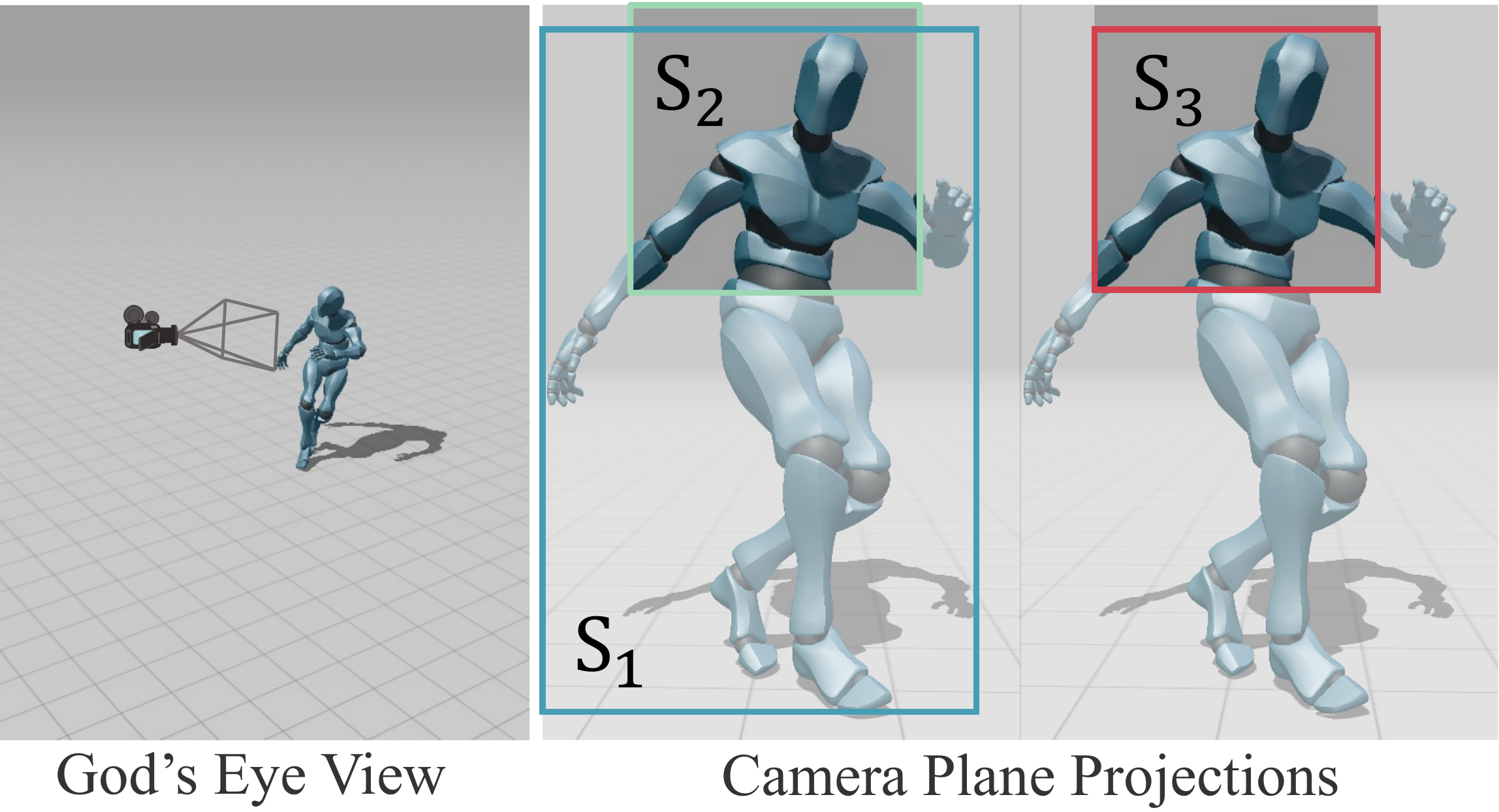}
  \caption{\textbf{Significant factors for shot features.} $S_1$ is the area of the dancer projected onto the camera plane, $S_2$ is the camera screen area and $S_3$ is the area of the dancer's body parts inside the camera screen. Here we use the character model from Mixamo~\cite{Mixamo}.
  }
  \label{fig:shotfeature}
  \vspace{-0.5cm}
\end{figure}
\subsection{Evaluation Metrics}
\textbf{Kinetic Feature Evaluation.} Following prior works~\cite{li2021ai,siyao2022bailando}, we evaluate generated camera movement using Frechet Inception Distance (FID) for quality and average Euclidean distance (Dist) in the feature space for diversity. For kinetic evaluation, we use a kinetic feature extractor~\cite{onuma2008fmdistance} following existing works~\cite{li2021ai,siyao2022bailando}. Since this feature extractor calculates average velocity and acceleration, we compute kinetic features on split 2.5-second data to ensure the density of feature distribution which is similar to settings in AIST++~\cite{li2021ai}. Thus, we have got FID$_k$ for kinetic quality and Dist$_k$ for kinetic diversity.\\
\textbf{Shot Feature Evaluation.} Shot features are significant to dance camera synthesis, however existing works~\cite{rao2020unified,tsingalis2012svm,canini2013classifying} are limited to 2D classifications with finite shot types. So we newly devise a shot feature extractor in 3D scenes, considering the knowledge of cinematography. As shown in Figure~\ref{fig:shotfeature}, we calculate shot features as:
\begin{equation}
{\rm Features}_{shot} = (S_3/S_1, S_3/S_2).
\end{equation}
where $S_1$ and $S_3$ indicate camera plane projection areas of the dancer's whole body and body parts inside the camera screen. $S_2$ is the camera screen area. Using this formulation, $S_3/S_1$ represents the percentage of the body inside the camera view and $S_3/S_2$ denotes the proportion of the camera screen that the dancer occupies. Then, we calculate FID and Dist for ${\rm Features}_{shot}$ and its velocity to get FID$_s$ and Dist$_s$ for shot quality and diversity. Considering the difference between shot and kinetic features, we compute shot metrics frame-by-frame to keep the accuracy of shot types.\\
\textbf{Dancer Fidelity Evaluation.} Dancer fidelity means camera movement should try to capture significant body parts against the dancer's poses and avoid the long time absence of the dancer in the camera view. We propose to evaluate dancer fidelity with the following two metrics: 1) Dancer Missing Rate (DMR): DMR represents the ratio of frames in which the dancer is not in the view of the camera, and 2) Limbs Capture Difference (LCD): LCD denotes the difference of body parts inside and outside camera view between synthesized results and ground truth. Lower DMR and LCD mean better dancer fidelity for fewer dancer-missing situations and more similarity between results and ground truth which is carefully modified. \\
\textbf{User Study.}
For qualitative evaluation, we conduct a user study to compare our method with baseline methods, ablation method, and ground truth. In this study, we first randomly select 10 dance-camera inputs from the test set ranging from 17$\sim$35 seconds and sample results from our methods and compared methods. For each baseline result, we combine the corresponding results from our method and produce 40 pairs of dance videos. Then, we invite 21 participants to watch all these 40 pairs of videos in a random shuffled order and answer the question “Which camera movement better showcases the dance and music? LEFT or RIGHT” for each pair of videos. The invited 21 participants consist of dancers, animators, filmmakers, and people who have rare expertise with camera and dance.

\begin{figure*}[h]
  \includegraphics[width=\linewidth]{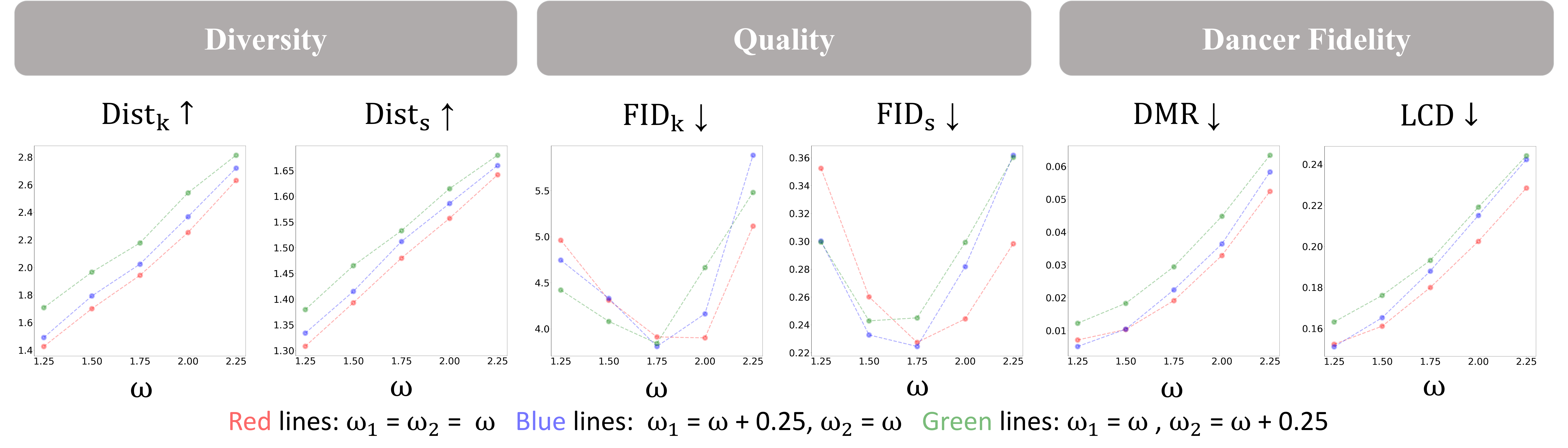}
  \vspace{-7mm}
  \caption{\textbf{Comparison of condition separation strategy on CFG.} \textcolor[RGB]{255, 96, 98}{Red} lines show the results of applying equal guidance weights $\omega_1,\omega_2$. Based on this, we separately add 0.25 guidance weight on $\omega_1$ and $\omega_2$, indicating enhancements in dance and music conditions which are represented with \textcolor[RGB]{108,104,249}{Blue} and \textcolor[RGB]{107,180,114}{Green} lines. Overall, CFG strengthens the diversity and quality of camera movement by trading off dancer fidelity. Compared to equal guidance on all conditions, adding guidance separately allows more fine-grained control of the trade-offs.
  }
  \label{fig:cfg}
  \vspace{-4mm}
\end{figure*}

\subsection{Comparison with Spatio-Temporal Models}
\begin{figure}[h]
  \includegraphics[width=\linewidth]{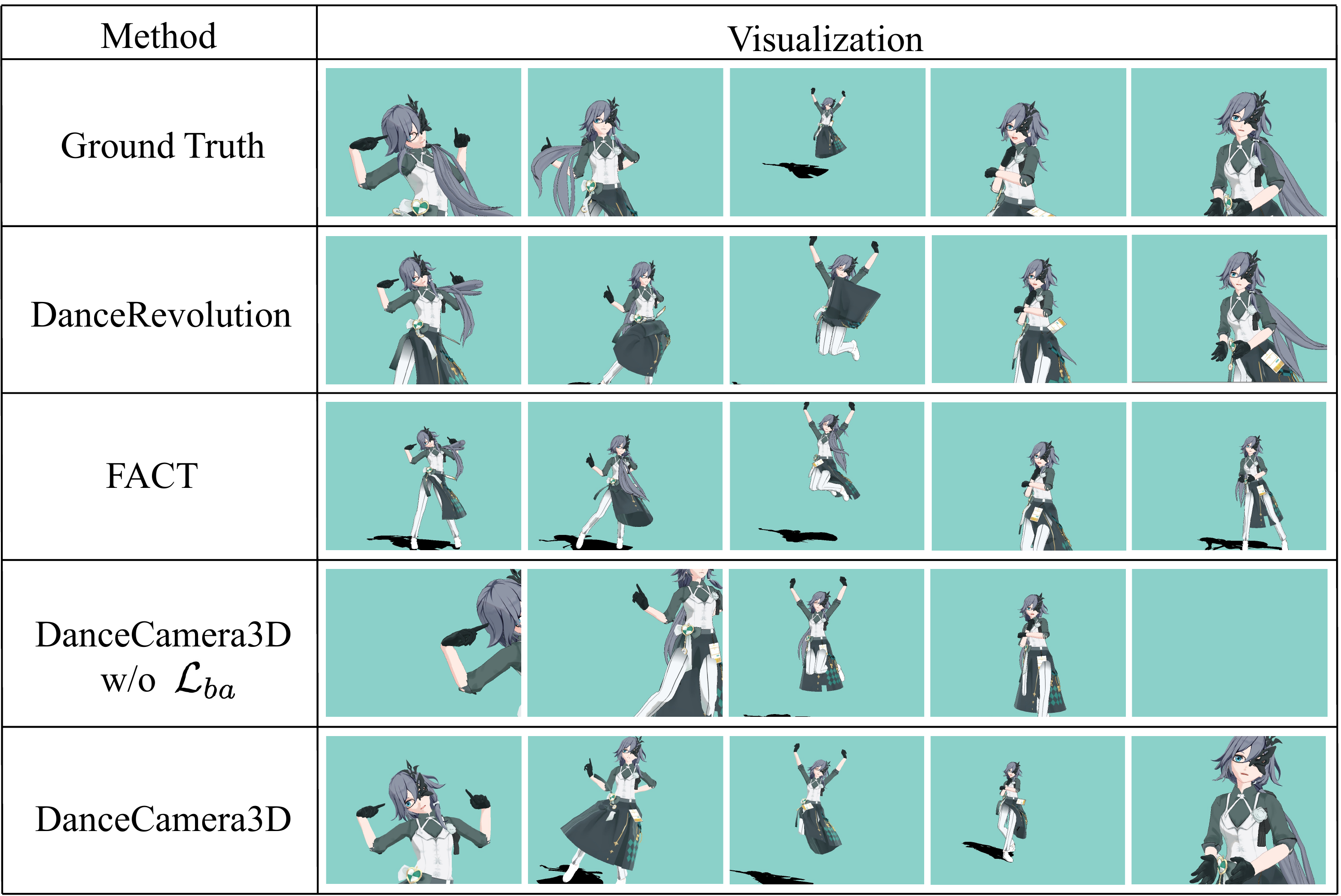}
  \caption{Visual comparison of rendered dance videos with synthesized camera movement from our DanceCamera3D and baseline methods. Compared to DanceRevolution and FACT, our DanceCamera3D produces more shot-type changes. DanceCamera3D w/o $\mathcal{L}_{ba}$ produces unbearable artifacts of failing to capture the position of the dancer which proves the effectiveness of $\mathcal{L}_{ba}$. Here we use the character model from \cite{Fuhua}.
  }
  \label{fig:case_study}
  \vspace{-0.5cm}
\end{figure}

Since there is no existing method for music-dance conditioned camera synthesis, we implement some baselines following an auto-regressive generation scheme which has achieved strong qualitative performance in dance synthesis:
\begin{itemize}
\item \textbf{DanceRevolution~\cite{huang2021}.} Following DanceRevolution, we synthesize camera movement autoregressively with a 3-layer LSTM decoder.
\item \textbf{FACT~\cite{li2021ai}.} Following FACT, we use a  transformer decoder to generate camera movement with 2.5-second seed movements in an autoregressive scheme. Bailando~\cite{siyao2022bailando} also achieves strong qualitative performance using transformers, but they pre-train dance motions as motion units which cannot be applied to the camera since camera movements have a strong correlation with dancers' positions so it's hard to encode them as independent units.
\end{itemize}
For comparison fairness, we utilize the same feature encoders and losses with our model. Results are shown in Table~\ref{tab:comparsononbackbone}, which demonstrate that our DanceCamere3D achieves better quality and higher diversity on both kinetic and shot features while preserving more dancer fidelity. The user study shows our method surpasses baseline methods by at least 65.71\% winning rate. Compared to ground truth camera movements, our synthesized camera movements still have 37.62\%. Feedback from users tells us that our model produces satisfying camera movements with considerable shot-type changes and a quite good focus on the character, but ground truth movements have more granular control and fewer artifacts since they are manually edited by animators. The case study also shows that our DanceCamera3D surpasses baseline methods and produces competitive results against ground truth, as illustrated in Figure~\ref{fig:case_study}.

\subsection{Comparison on CFG}\label{sec:cmpcfg}
Classifier-free guidance (CFG) has been demonstrated to achieve state-of-the-art results for image generation~\cite{ho2022classifier, sun2023inner} and dance synthesis~\cite{tseng2023edge} using explicit control over the diversity-fidelity trade-off. However, dance camera synthesis is a more complex situation, since this problem has 2 conditions in which dance motion is strongly correlated to the camera and music audio is weakly correlated. Thus, we devise a strategy to separate dance and music conditions and conduct experiments on it. As shown in Figure~\ref{fig:cfg}, we illustrate results with different guidance weights $\omega_1,\omega_2$ to dance and music conditions including: 1) Red lines: $\omega_1 = \omega_2 = \omega$, 2) Blue lines: $\omega_1 =\omega + 0.25$, $\omega_2 = \omega$ and 3) Green lines: $\omega_1 =\omega$, $\omega_2 = \omega + 0.25$. Overall, results demonstrate that CFG strengthens the diversity and quality of camera movement while trading off dancer fidelity. Notably, too large guidance weights cause a drop in the quality of camera movement. This is because overdose enhancement on conditions will move the results away from the ground truth distribution. Comparing green and blue lines, we find the music condition produces a more intense effect on camera movements, and the dance motion condition causes slower changes and better quality for achieving lower FID$_k$ and FID$_s$ with less loss on dancer fidelity at some points. This complies with the reality that, dance motion as a strong condition brings more focus on dancers, and music as a weak condition influences more on movement style. In summary, our strong-weak conditions separation strategy provides more fine-grained control on the trade-offs in dance camera synthesis. 
% \vspace{-5mm}
% \vspace{-3mm}
\subsection{Ablation Study}\label{sec:ablation}
We conduct an ablation experiment to study the effectiveness of our newly designed body attention loss $\mathcal{L}_{ba}$. As shown in Table~\ref{tab:comparsononbackbone}, quantitative evaluations show that the quality, dancer fidelity, and kinetic diversity decrease when we remove the $\mathcal{L}_{ba}$. The diversity of shot increases since there are more frames without dancer in the camera screen, which greatly change the distribution of shot features. User study also proves that model with $\mathcal{L}_{ba}$ produces more stable results with fewer artifacts like failing to capture the dancer or placing the dancer at the edge of the screen for a long time. As shown in Figure~\ref{fig:case_study}, the model without $\mathcal{L}_{ba}$ is more likely to generate unbearable artifacts which demonstrate the effectiveness of $\mathcal{L}_{ba}$.

% \vspace{-5mm}
\vspace{-3mm}
\section{Conclusion and Future Work}
\label{sec:conclusion}

In this paper, we introduce a novel and valuable task: Music Dance driven Camera Movement Synthesis. To address this challenging problem, we constructed a new dataset DCM, which for the first time simultaneously collects camera, dance, and music data together with rich annotations.  With this dataset, we present DanceCamera3D with a novel loss function and condition separation strategy, that can synthesize high-quality 3D dance camera movement given music and dance. We conduct comprehensive evaluations on the DCM dataset with newly devised metrics. Through extensive experiments, we demonstrate the effectiveness of our DanceCamera3D. To encourage further research in the fields of music, choreography, and cinematography, we will make both the source code and the dataset openly available as open-source resources. We believe our DCM dataset can not only facilitate the studies of dance camera synthesis but also contribute to research like camera keyframing and shot type classification.
\section{Acknowledgements}
This work is supported by the National Key R\&D Program of China under Grant No.2021QY1500. \\

{
    \small
    \bibliographystyle{ieeenat_fullname}
    \bibliography{main}
}


\appendix
\section*{{\Large Appendix}}
\section{Calculation of Joint Mask}
\begin{table*}[h]
\setlength\tabcolsep{3pt}% 调整列间距
\vspace{-2mm}
  \begin{tabular}{cccccccccc}
    \toprule{}
    \multirow{2}{2cm}[-0.5ex]{\centering Music \\Languages}&\multirow{2}{1cm}[-0.5ex]{\centering Total\\ Time}& \multicolumn{2}{c}{BPM}& \multicolumn{2}{c}{Camera Keyframes Interval} &\multicolumn{2}{c}{Number of Sequences}&\multicolumn{2}{c}{Frames of Sequences}\\
    \cmidrule(lr){3-4} \cmidrule(lr){5-6} \cmidrule(lr){7-8} \cmidrule(lr){9-10}
    &&Range & Average&\quad Range & Average & \quad Aligned & Split& Aligned & Split\\
    \midrule
    Chinese &4298.8s&77.8$\sim$143.6 & 119.2&\quad 1$\sim$475&18.45&\quad 38&169&524$\sim$8025&510$\sim$1051\\
    Japanese & 2258.8s & 86.1$\sim$161.5&129.4&\quad 1$\sim$220&6.67&\quad15&88&1618$\sim$7290&512$\sim$1046\\
    Korean & 3996.5s&86.1$\sim$161.5&119.8&\quad1$\sim$231&11.93&\quad40&159&550$\sim$6260&516$\sim$1042\\
    English & 916.6s & 99.4$\sim$143.6&122.6&\quad1$\sim$228&18.57&\quad15&38&829$\sim$5737&510$\sim$1046\\
  \bottomrule{}
  
\end{tabular}
\caption{\textbf{Detailed statistics of the DCM dataset.} ‘Aligned’ means data after alignment among dance, camera, and music. ‘Split’ denotes data split into subsequences within 17$\sim$35s which is more suitable for training.}
\label{tab:detailed_statistics}
\end{table*}
\begin{figure}[h]
  \includegraphics[width=\linewidth]{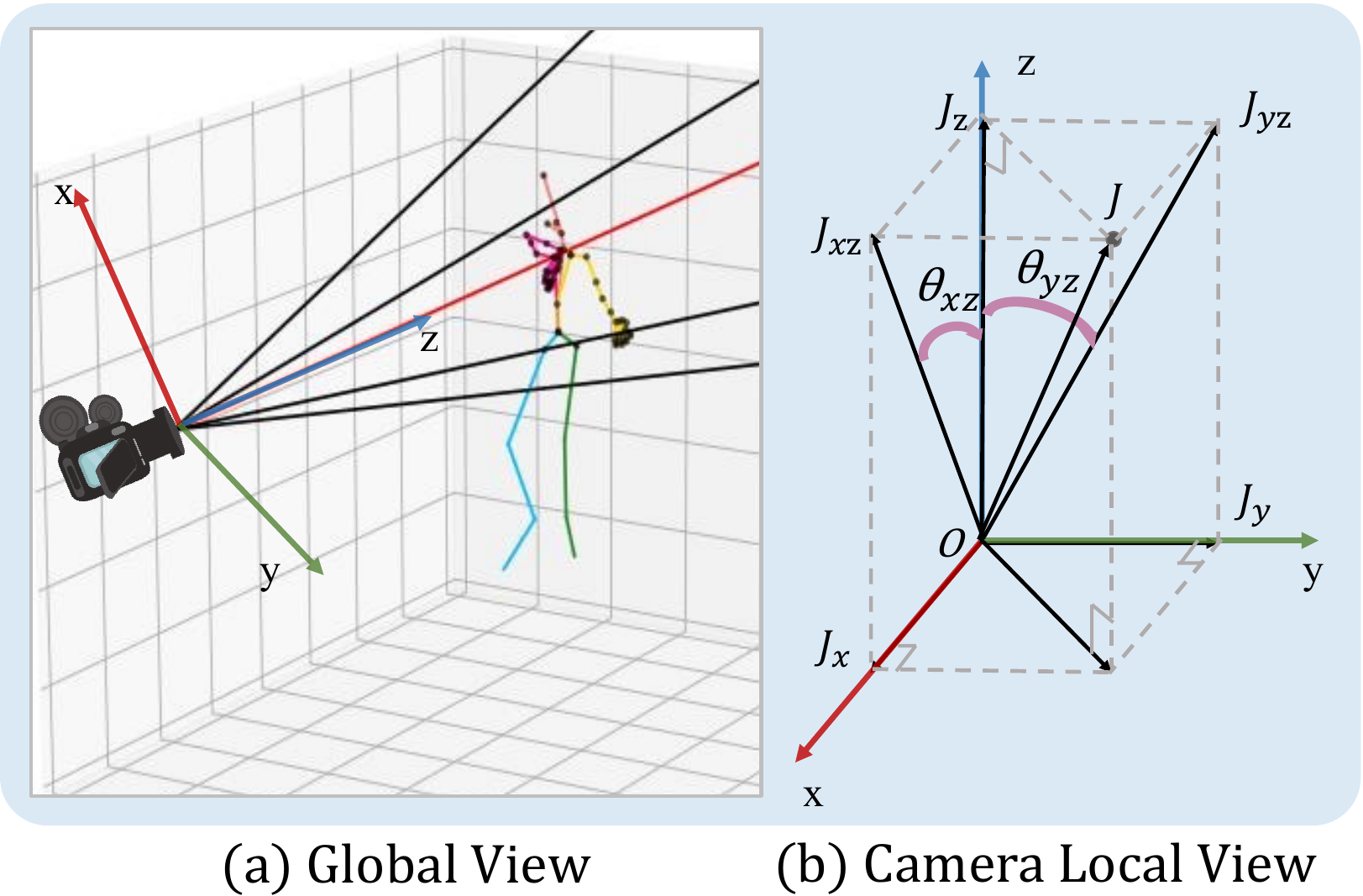}
  \caption{\textbf{Details of joint mask calculation.}
  }
  \label{fig:joint_mask}
\end{figure}
As shown in Figure~\ref{fig:joint_mask}, we illustrate camera coordinates in (a) Global View and (b) Camera Local View. Given that $O$ is the camera eye at frame $i$ and $J$ is an arbitrary joint at frame $i$, our target is to determine whether $J$ is inside the camera view or not. Specifically, we first project $J$ onto the $xz$-plane and the $yz$-plane to get $J_{xz}$ and $J_{yz}$, as shown in (b) of Figure~\ref{fig:joint_mask}. Supposing that joint masks of joints inside camera view are 1 and others are 0, we can represent the joint mask of $J$ as $Jm$:
\begin{equation}
Jm = 
\begin{cases}
1& \theta_{xz} \leq Fov/2, \theta_{yz} \leq Fov/2, \\
0& \text{others},
\end{cases}
\end{equation}
where $\theta_{xz}$ is the angle between $\overrightarrow{OJ}_{xz}$ and $\overrightarrow{OJ}_{z}$, $\theta_{yz}$ is the angle between $\overrightarrow{OJ}_{yz}$ and $\overrightarrow{OJ}_{z}$, and $Fov$ is the field of camera view at frame $i$. In more detail, we use the cosine function to compare angles, because the cosine function is symmetrical about 0 and monotonically decreases on $[0,\pi)$. Thus, we can further represent $Jm$ as:
\begin{equation}
Jm = 
\begin{footnotesize}
\begin{cases}
1&  Cos(\theta_{xz}) \geq Cos(Fov/2), Cos(\theta_{yz}) \geq Cos(Fov/2), \\
0& \text{others},
\end{cases}
\end{footnotesize}
\end{equation}
For computing $Cos(\theta_{xz})$ and $Cos(\theta_{yz})$, we take $Cos(\theta_{xz})$ as an example:

\begin{equation}
\begin{split}
	Cos(\theta_{xz}) &= \frac{\overrightarrow{OJ}_{xz}*\overrightarrow{z_0}}{||\overrightarrow{OJ}_{xz}||*||\overrightarrow{z_0}||},\\
        \overrightarrow{OJ}_{xz} &= \overrightarrow{OJ} - \overrightarrow{OJ}_{y},\\
	\overrightarrow{OJ}_{y} &= (\overrightarrow{OJ}*\overrightarrow{y_0})\overrightarrow{y_0},
\end{split}
\end{equation}

where $\overrightarrow{x_0}, \overrightarrow{y_0}, \overrightarrow{z_0}$ are unit vectors of $x, y, z$ axes. Here $\overrightarrow{x_0}, \overrightarrow{y_0}, \overrightarrow{z_0}$ and the position of $O$ make up the camera-centric format representation $xc$, which is mentioned in Sec \textcolor[RGB]{255, 0, 0}{3.2} and Sec \textcolor[RGB]{255, 0, 0}{4.1} of the full paper.

\begin{figure}[h]
  \includegraphics[width=\linewidth]{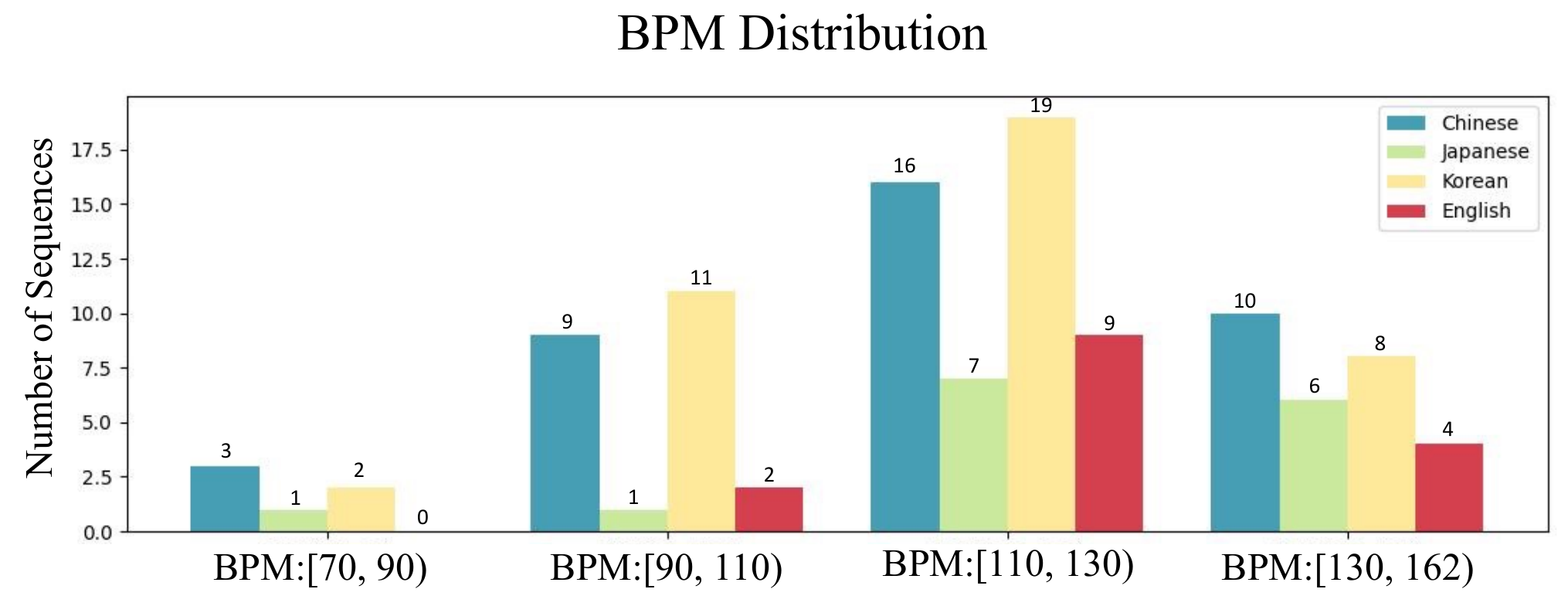}
  \caption{\textbf{BPM Distribution of DCM Dataset.}
  }
  \label{fig:bpm_distribution}
\end{figure}

\section{Implementation of Body Attention Loss}
In Sec \textcolor[RGB]{255, 0, 0}{4.3}, we represent our body attention loss $\mathcal{L}_{ba}$ as:
\begin{equation}
\mathcal{L}_{ba} = ||\boldsymbol{Jm} - \boldsymbol{\hat{J}m}*\boldsymbol{Jm}||, 
\label{eq:oldlba}
\end{equation}
where $\boldsymbol{\hat{J}m}$ means the generated joint mask and $\boldsymbol{Jm}$ means the ground-truth joint mask. This concise and clear representation denotes that we penalize the joints that are inside the camera view in ground truth but outside the camera view in synthesized results. However, in the actual implementation of $\mathcal{L}_{ba}$, we find that the calculation of joint mask is underivable. Thus, we implement $\mathcal{L}_{ba}$ as:
\begin{equation}
\begin{aligned}
\mathcal{L}_{ba} &= {\rm Relu}(\boldsymbol{Jm} * ({\rm Cos}(\frac{Fov}{2}) - {\rm Cos}(\boldsymbol{\theta_{xz}}))) \\
&+{\rm Relu}(\boldsymbol{Jm} * ({\rm Cos}(\frac{Fov}{2}) - {\rm Cos}(\boldsymbol{\theta_{yz}})))
\end{aligned}
\end{equation}
where $\frac{Fov}{2}$ indicates the vector of camera field of view, $\boldsymbol{\theta_{xz}}$ and $\boldsymbol{\theta_{yz}}$ denote vectors of $\theta_{xz}$ and $\theta_{yz}$ respectively. In this way, for the joint outside the camera field of view in ground truth, the $Jm$ is 0 and the corresponding impact to $\mathcal{L}_{ba}$ is 0. For the joint inside the camera field of view in ground truth, the $Jm$ is 1, and the corresponding impact to $\mathcal{L}_{ba}$ is 0 only if this joint is inside the camera field of view in the generated result. Thus, this loss realizes a similar penalty as Equation~\ref{eq:oldlba}.

\section{Details of DCM Dataset}

\subsection{BPM}
The BPMs (beat per minute) of music pieces in our DCM dataset range from 71 to 162. In detail, we illustrate BPM distribution with music categories in Figure~\ref{fig:bpm_distribution}.

\subsection{Detailed Statistics}
As shown in Table~\ref{tab:detailed_statistics}, we present more detailed statistics of our DCM dataset.

\end{document}